\documentclass{article}
\usepackage{ijcai18}

\usepackage{times}
\usepackage{xcolor}
\usepackage{soul}
\usepackage[utf8]{inputenc}
\usepackage[small]{caption}
\usepackage{epsfig}
\usepackage{graphicx}
\usepackage{amsmath}
\usepackage{amssymb}
\usepackage{graphics,picture}
\usepackage{enumerate}
\usepackage{multicol}
\usepackage{multirow}

\newcommand{\work}[1]{\textcolor{black}{#1}}
\newcommand{\wyb}[1]{\textcolor{black}{#1}}
\newcommand{\hb}[1]{\textcolor{black}{#1}}
\newcommand{\yuanbo}[1]{\textcolor{black}{#1}}

\title{Active Object Reconstruction Using a Guided View Planner}

\author{
Xin Yang\protect\footnotemark[1]$^1$~$^2$,
Yuanbo Wang\protect\footnotemark[1]$^1$,
Yaru Wang$^1$,
Baocai Yin$^1$,
Qiang Zhang$^1$
Xiaopeng Wei$^1$,
Hongbo Fu$^2$
\\
$^1$ Dalian University of Technology~~~~~~~~~~~~~~~$^2$ City University of Hong Kong\\
xinyang@dlut.edu.cn, yuanbodlut@gmail.com, wangyaru@mail.dlut.edu.cn
\\
\{ybc, zhangq, xpwei\}@dlut.edu.cn, hongbofu@cityu.edu.hk
}
\begin{document}

\maketitle

\begin{abstract}
Inspired by the recent advance of image-based object reconstruction using deep learning, we present an active reconstruction model using a guided view planner. We aim to reconstruct a 3D model using images observed from a planned sequence of informative and discriminative views. But where are such informative and discriminative views around an object? To address this we propose a unified model for view planning and object reconstruction, which is utilized to learn a guided information acquisition model and to aggregate information from a sequence of images for reconstruction. Experiments show that our model (1) increases our reconstruction accuracy with an increasing number of views (2) and generally predicts a more informative sequence of views for object reconstruction compared to other alternative methods.
\end{abstract}

%----------------------------------------------------------------------------------
\protect\footnotetext[1]{Equal Contribution.}
%----------------------------------------------------------------------------------

%------------------------------------------------------------------------
\begin{figure*}[tbp]
  \begin{minipage}{0.33\linewidth}
  \centerline{\includegraphics[width=1\textwidth]{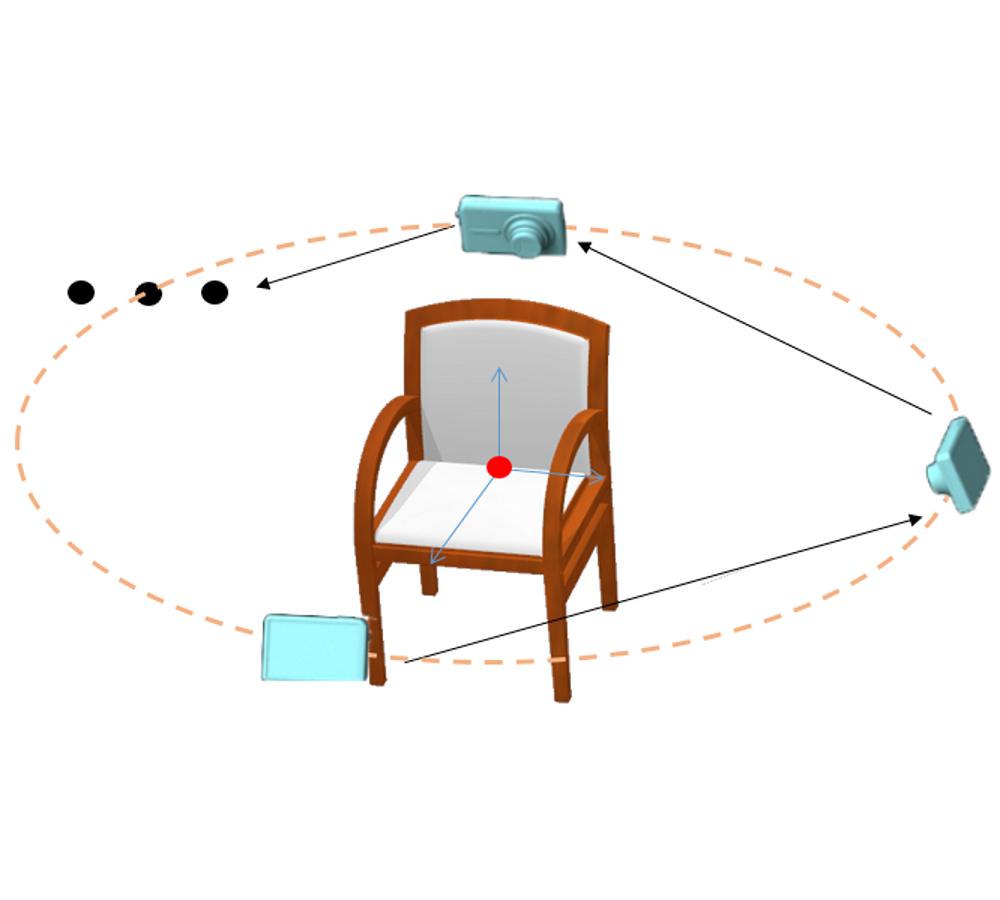}}
  \vspace{1.5ex}
  \centerline{(a) Viewing Space}
  \vspace{1ex}
  \end{minipage}
  \begin{minipage}{0.66\linewidth}
  \centerline{\includegraphics[width=1\textwidth]{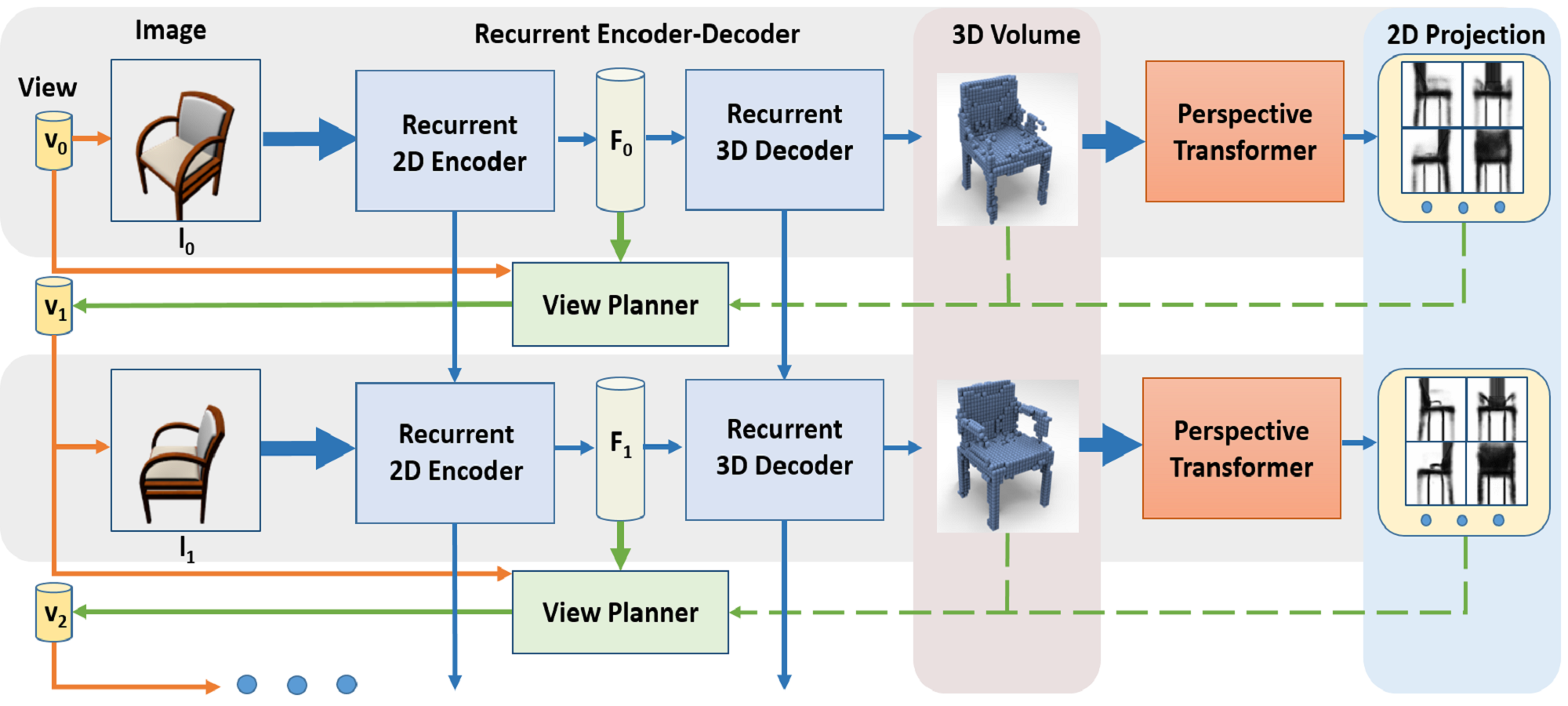}}
  \vspace{1.5ex}
 \centerline{(b) Active Object Reconstruction Model}
 \vspace{1ex}
  \end{minipage}
  \caption{\label{fig:fig1}We present an active object reconstruction model. Targeting a 3D object, we utilize the guidance from both 3D volume and 2D projection to train a View Planner, which continuously predicts the next informative and discriminative views parameterized as camera azimuth angles on a viewing circle \protect\footnotemark[2] around the object (a). Based on the predicted view sequence, our Recurrent Encoder-Decoder takes the feature sequence encoded by a Recurrent 2D Encoder as input and recurrently decodes it to a 3D volume by using a Recurrent 3D Decoder (b), which gradually improves the accuracy with an increasing number of views.}
\end{figure*}
%------------------------------------------------------------------------
\section{Introduction}
\label{section:section1}

With the growing application demand of robot-object manipulation and 3D printing, automatic and efficient 3D model reconstruction from 2D images has recently been a hot topic in the research field of computer vision. Classic 3D reconstruction methods, based on the Structure-from-Motion technology ~\cite{halber2016fine,snavely2006photo}, are usually limited to the illumination condition, surface textures and dense views. On the contrary, benefited from prior knowledge, learning-based methods \yuanbo{~\cite{liu2017stochastic,yan2016perspective}} can utilize a small number of images to reconstruct a plausible result without the assumptions on the object reflection and surface textures.

Regarding object reconstruction as a predictive and generative issue from a single image, learning based methods usually utilize CNN-based encoder and decoder to predict a 3D volume ~\cite{girdhar2016learning,dai2016shape,wu2016learning,kar2015category-specific}, 3D structure ~\cite{wu2016single} or point set ~\cite{fan2016point} trained by 3D supervision. Recent work ~\cite{yan2016perspective,tulsiani17multi} attempts to use only 2D images \work{for} 2D supervision to train image-based object reconstruction \work{models}. \hb{For example}, with a differentiable style, \hb{Yan et al.~\shortcite{yan2016perspective}} propose Perspective Transformer Nets with a novel projection loss that enables the 3D learning using 2D projection without 3D supervision. Tulsiani et al.~\shortcite{tulsiani17multi} study the consistence between a 3D shape and 2D observations and propose a differentiable formulation to train 3D prediction by 2D observations.

A crucial assumption in the above-mentioned models, however, is that the input images contain most information of a 3D object. As a result, these models fail to make a reasonable prediction when the observation has severe self-occlusion as they lack the information from other views. An effective solution is to utilize more views to make up the information. Choy et al.~\shortcite{choy20163d} propose a 3D Recurrent Neural Networks (\emph{\textbf{3D-R2N2}}) to map multiple random views to their underlying 3D shapes. In contrast to 3D-R2N2, we focus on how much information is sufficient and how to aggregate these information for 3D object reconstruction. In other words, how many views and which views of image can capture the most informative feature and maximize the quality of reconstruction? This is a problem about dynamical view prediction when reconstructing an object. It means an active process of capturing new information for 3D reconstruction.

There are many methods receiving the maximal information gain such as the decrease of uncertainty ~\cite{xu2015autoscanning}, Monte Carlo sampling ~\cite{denzler2002information} and Gaussian Process Regression ~\cite{huber2012bayesian}. Some methods regard this information gain task as a sequential prediction of the next best view (NBV) by reducing scanning effort ~\cite{wu2014quality} or uncertainty of object ~\cite{wu20153d} with the least observations. All these attempt to receive the maximal information gain with the minimal number of views. Intuitively, it is not absolute that the best performance comes from the maximal information. The reason is perhaps that learning based methods attempt to exploit the spatial and temporal structure of the sequential observations ~\cite{xu20163d} and learn to predict an approximate views sequence to optimize the deviation between the prediction and the ground truth. The attention model ~\cite{mnih2014recurrent} is a good solution for the problem of sequential locations prediction as it utilizes the recurrent neural network to extract the information from a single image or an image sequence, and adaptively selects a sequence of discriminative regions or locations. There are many successful applications using the attention model such as image classification ~\cite{xiao2015application}, image captioning ~\cite{xu2015show} and 3D shape recognition ~\cite{xu20163d}.

However, \hb{different from image identification building the inconsistency between predicted category labels and ground truth, object reconstruction requires a dense prediction for each voxel}, and it thus needs to explore a deeper relation between 3D volume and 2D images and to use this relation to guide the aggregation of \hb{multi-view information} and the planning of sequential views. To achieve this, our method differs from \hb{the other attention-based models} in two major aspects. First, to constrain the consistency between 3D volume and 2D images, we combine the volumetric and projective supervision in the process of view aggregation. Second, for guided view planning, our reward is set upon the performance of reconstruction and volume-projection consistency, facilitating the view planner to capture more information.
%----------------------------------------------------------------------------------
\protect\footnotetext[2]{Note that our model can apply to a viewing sphere as well but we found viewing parameters in a circle are enough for the training of the synthesis data.}
%----------------------------------------------------------------------------------

Our experiments show that our model aggregates more discriminative information from multi-view images and apparently increases the accuracy with an increasing number of views. For the view planning task, we demonstrate our sequences can \work{give better prediction} than other strategies in the test. The main contributions of this paper are as follows: (1) We build a \wyb{Recurrent Encoder-Decoder} based on multiple Conv-RNN layers and a volume-projection supervision, \hb{leading to a better reconstruction performance.} (2) We combine 3D volume prediction and 2D projection to design the reward for view planning policy learning. Under the control of the combined reward, we can implicitly learn the deep relationship between 3D reconstruction and 2D images, and optimize the planning policy. (3) We \wyb{propose} an active framework that learns a view planner for 3D object reconstruction. Our model can dynamically determine view selection based on information gain and discrimination, which makes the reconstruction \hb{more accurate}.

%------------------------------------------------------------------------
\begin{figure*}[tp]
  \centering
  \includegraphics[width=0.99\linewidth]{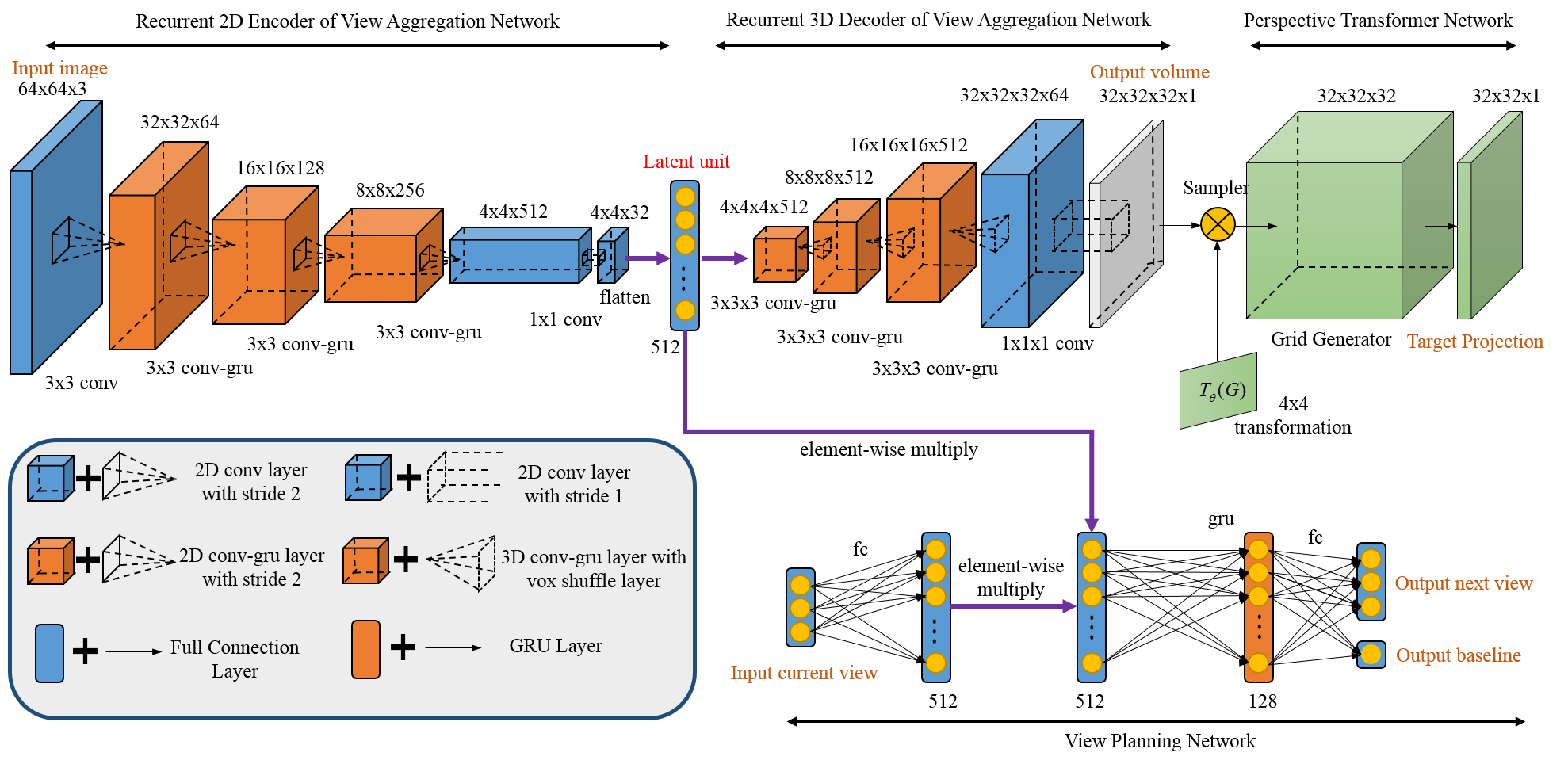}
  \caption{\label{fig:fig2}
          Illustration of network architecture. Our entire network consists of four components: \wyb{a Recurrent 2D Encoder, a Recurrent 3D Decoder, Perspective Transformer and a View Planner}. Taking current view and a rendered RGB image as input, the network predicts occupancy probability for each voxel in a $32\times32\times32$ volume and the next view.}
\end{figure*}
%----------------------------------------------------------------------------------

\section{Methodology}
\subsection{Overview}
\label{section:section2.1}

 Figure \ref{fig:fig1} illustrates our model, which is summarized as follows. \hb{In the training stage, starting at a random view in the viewing circle (Figure \ref{fig:fig1} (a)), we select the pre-rendered color image with its associated view close to the random view} and feed it into a \textbf{Recurrent 2D Encoder}, which encodes the image to a latent unit $F_{0}$ and propagates the information when absorbing new views. The encoded unit is then fed into two branch pathways. One is a \textbf{Recurrent 3D Decoder}, which maps the latent unit extracted from all past views to a predicted 3D volume. The other one called \textbf{View Planner} serves as a dynamical view prediction module that continually receives and integrates the encoded units from current and all past views (Figure \ref{fig:fig1} (b)), and regresses view parameters for next observation (Section \ref{section:section2.2}). To combine view planning and object reconstruction into a unified and correlative model, we propose a volume-projection guidance (Section \ref{section:section2.3}) for the supervised learning of view-based volume mapping and reinforcement learning of continuous view prediction. The volume comes from the Recurrent Encoder-Decoder and the projection is generated by a differentiable \textbf{Perspective Transformer} ~\cite{yan2016perspective}. \hb{In the test stage, our model keeps acquiring the images observed from a target object under the guidance from the View Planner and reconstructs the 3D model with the Recurrent Encoder-Decoder.}

\subsection{Network Architecture}
\label{section:section2.2}
\wyb{We consider multi-view volumetric reconstruction as a dense prediction from a sequence of views and develop a unified framework for both volume reconstruction and view planning in an active scheme. The procedure can be divided into multiple time steps and we plot one step of data flow in Figure \ref{fig:fig2}. Next we discuss the detailed architecture.}

\textbf{Recurrent 2D Encoder.} This network is utilized to extract features from input image $I_{t}$ and aggregate them with the past views to a latent unit: $F_{t}=f_{enc}(I_{t},S^{enc}_{t-1})$, where $S^{enc}_{t-1}$ refers to all past states of hidden layers in the encoder network. In our implementation, we build a Recurrent 2D Encoder upon multiple 2D Conv-GRU layers to extract the spatial features from images and integrate the sequential past states. 2D Conv-GRU layers update their hidden states under the control of three convolution-operator gates with the hidden states arranged in 2D space. Compared to convolutional layers based auto-encoder networks with a single 3D-GRU, our network is better at \hb{feature extraction} on image sequence, since our reconstruction performance improves a large margin (demonstrated in Section \ref{section:section3.4}).

\textbf{Recurrent 3D Decoder.} Taking features $F_{t}$ extracted by the encoder network as input, Recurrent 3D Decoder utilizes multiple 3D Conv-GRU layers, which are similar to 2D Conv-GRU layers but arranged in 3D space, to decode $F_{t}$ to a 3D volume where each voxel grid retains the probability of occupancy: $V_{t}=f_{dec}(F_{t},S^{dec}_{t-1})$, where $S^{dec}_{t-1}$ refers to all past states of hidden layers in the decoder network. To increase the resolution of feature maps, we add a voxel shuffle layer after each 3D Conv-GRU layer, which allocates the depth dimension of feature vector to 3D space.

\textbf{View Planner.} The task of view planning is to actively regress a sequence of views parameterized as camera azimuth angles on a viewing circle around the object. Taking a random angle as initial view, we sequencely feed the rendered image under the current view into the Recurrent 2D Encoder to extract and aggregate features. We then merge the features with the current view parameters by element-wise multiplication to get a viewing ``glimpse'' \cite{mnih2014recurrent}, which fuses the information of both the image sequence and current view: $ g_t=f_{enc}(I_{t},S^{enc}_{t-1})*f_{view}(v_{t-1})$. With a GRU layer, our model retains all past glimpse information and continuously absorbs new views. The glimpse information can be formulated as $h_t^{gru}=f_{gru}(g_t,h_{t-1}^{gru})$, which discriminatively describes the relation of images sequence, 3D volume and parameters of a sequence of views. Feeding the states to an extra full connection layer after the GRU layer, we finally predict view parameters of the next view to get the next image input.

\textbf{Perspective Transformer.} We use the Perspective Transformer Network proposed by Yan et al.~\shortcite{yan2016perspective} to obtain a 2D projection from the 3D volume. Utilizing this 3D differentiable transformation, we project the 3D voxel prediction to a \work{2D grid, which} looks like a projection silhouette. Combining this differentiable 2D projection with the predicted volume, we build a projective guidance on the training of volume prediction and view planning (see Section \ref{section:section2.3} for details).

\subsection{Volume-Projection Guidance}
\label{section:section2.3}

\wyb{We combine the procedure of object reconstruction and view planning into a unified framework. For view planning, it optimizes a view prediction policy under the control of feedback signals based on the evaluation of reconstruction performance (as shown in Figure \ref{fig:fig1} (b)). For object reconstruction, the Recurrent Encoder-Decoder receives input images from the informative views predicted by the View Planner, which ensures a sufficient information gain and boosts the improvement of reconstruction performance. We jointly train two modules by both volumetric and projective patterns but use different strategies: a reinforcement learning under the control of volume-projection reward and a supervised learning using volume-projection supervision.}

\textbf{Volume-Projection Reward.} At each time step, the guidance of View Planner comes from the performance of volume predicted by the reconstruction module. In other words, the View Planner receives a reward signal which is built upon the reconstruction feedback from the Recurrent Encoder-Decoder. We only calculate the accumulative reward during a whole episode to update a view planning policy, which maps the image observation to the camera view. \hb{To accommodate the dense prediction on the whole 3D volume} and ensure the reconstruction improvement with new views fed in, the increment of voxel Intersection-over-Union (IoU) is utilized to measure the reconstruction reward. Mathematically, the reward at step t can be formulated as follows:
\begin{equation}\label{eqn1}
    r^{t}_{cons}=IoU(\widehat{V}_{t},V)-IoU(\widehat{V}_{t-1},V),
\end{equation}
where $\widehat{V}_{t}$ is the 3D volume predicted by the Recurrent Encoder-Decoder, and $V$ is the corresponding ground truth in the dataset.

To implicitly learn the relation between 3D volume and 2D projection, we design a projection reward to encourage the consistence between 3D construction and 2D projection from different \work{viewpoints}. The projection reward is defined as the increment of pixel IoU value on 2D silhouettes sampled by the Perspective Transformer from \work{multiple} different views. The reward at step t can be formulated as follows:
\begin{equation}\label{eqn2}
\begin{split}
  r^{t}_{proj}=\frac{1}{n}\sum^{n}_{i=1}&IoU(f_{ptn}(\widehat{V}_{t},v_{i}),f_{ptn}(V,v_{i}))\\ -&IoU(f_{ptn}(\widehat{V}_{t-1},v_{i}),f_{ptn}(V,v_{i})),
\end{split}
\end{equation}
\hb{where $n$ is the number of projection views and $v_i$ is the $i-th$ view.}

In addition, to punish for selecting similar views, we add an additional movement cost defined as the minimum value of the circle distance between the current location and past views. Integrating the reconstruction reward, projection reward, and movement cost, the final reward is defined as follows:
\begin{equation}\label{eqn3}
  r=\lambda_{v}r_{cons}+\lambda_{p}r_{proj}-\lambda_{m}C_{move},
\end{equation}
where $\lambda_{v}$, $\lambda_{p}$ and $\lambda_{m}$ are the weights of the reconstruction reward, the projection reward, and the movement cost, respectively.

Using this reward, we can control the update of policy, which corresponds to the gradient policy algorithm. We sample the views predicted by the View Planner according to a normal distribution with a predefined standard deviation at each time step, and minimize the following loss function to optimize the view planning policy:
\begin{equation}\label{eqn4}
 L_{rl}=\sum_{t=1}^{t=T}-log(p(v^{p}_{t}\mid I_{t},\theta_{vp}))*(R_{t}-b_{t}),
\end{equation}
where $v_t^{p}$ is \work{a} sampled view at time step $t$, $R_{t}$ is the reward at time t, and $b$ is a predicted value as a various baseline which is utilized to center the reward (\cite{mnih2014recurrent}). The log probability can derivative by back propagation of network and reward $R$ is the signal received from the reconstruction feedback of Recurrent Encoder-Decoder.

\textbf{Volume-Projection Supervision.} The loss function of Recurrent Encoder-Decoder is defined as the mean value of voxel-wise square error (MSE):
\begin{equation}\label{eqn5}
  L_{vox}=\|V_{pre}-V\|^{2},
\end{equation}
where $V_{pre}$ is the final output of the Recurrent Encoder-Decoder.

\wyb{Besides the 3D volumetric loss, we add a 2D projective loss to implicitly learn the effect of 2D projection on 3D prediction, which improves the multi-view reconstruction performance.} The 2D supervision loss is formulated as:

\begin{equation}\label{eqn6}
  L_{proj}=\frac{1}{n}\sum^{n}_{j=1}\|f_{ptn}(V_{pre},T^{j})-M^{j}\|^{2},
\end{equation}
where $f_{ptn}$ is the Perspective Transformer Network, $T^j$ is the parameters of $j-th$ view with a $4$-by-$4$ transformation matrix, and $M^j$ is the projection of the ground truth voxel. We combine the 3D supervision (Equation \ref{eqn5}) and 2D supervision (Equation \ref{eqn6}) using a weighted sum as:
\begin{equation}\label{eqn7}
  L=\lambda_{vox}L_{vox}+\lambda_{proj}L_{proj},
\end{equation}
where $\lambda_{vox}$ and $\lambda_{proj}$ are the weights of volumetric loss and perspective loss, respectively.

\section{Evaluation}

In this section, we discuss the following three questions: \hb{(1) Can our model improve the accuracy of reconstruction with an increasing number of views? (Section \ref{section:section3.2}) (2) Can our View Planner obtain more informative and discriminative views to boost the reconstruction performance compared to the other alternative methods?} (Section \ref{section:section3.3}) (3) Do our network structures learn better than other settings? (Section \ref{section:section3.4})

%----------------------------------------------------------------------------------
\begin{figure*}[tp]
\begin{minipage}[b]{0.29\linewidth}
\centering
\includegraphics[width=1\textwidth]{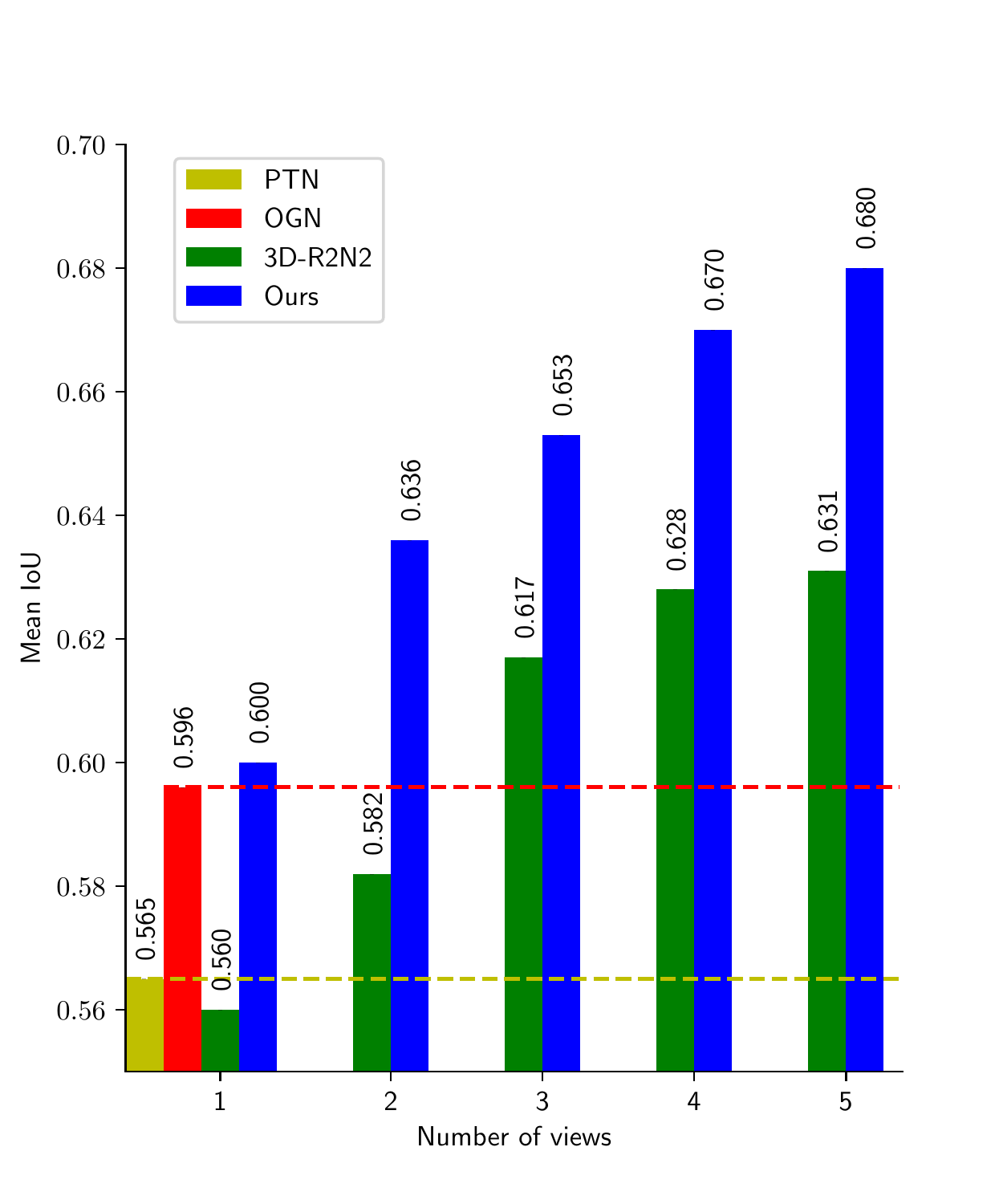}
\centerline{(a) Overall results}
\end{minipage}
\begin{minipage}[b]{0.7\linewidth}
\centering
\begin{tabular}{| c | c | c | c  c  c | c  c  c |}
\hline
{methods} &{PTN} &{OGN} &\multicolumn{3}{c|}{3D-R2N2} &\multicolumn{3}{c|}{Ours}\\
\hline
\# views &1 &1 &1 &3 &5 &1 &3 &5\\
\hline
plane     &0.553 &0.587 &0.513  &0.549  &0.561  &0.605  &0.657  &\textbf{0.679}  \\
bench    &0.482 &0.481 &0.421  &0.502  &0.527  &0.498  &0.569  &\textbf{0.597}  \\
cabinet   &0.711 &0.729 &0.716  &0.763  &0.772  &0.715  &0.769  &\textbf{0.789}  \\
car     &0.712  &0.816 &0.798  &0.829  &0.836  &0.757  &0.805  &\textbf{0.838}  \\
chair    &0.458  &0.483 &0.466  &0.533  &0.550  &0.532  &0.590  &\textbf{0.617}  \\
monitor     &0.535 &0.502 &0.468  &0.545  &0.565  &0.524  &0.596  &\textbf{0.624}  \\
lamp    &0.354  &0.398 &0.381  &0.415  &0.421  &0.415  &0.445  &\textbf{0.461}  \\
speaker    &0.586  &0.637 &0.662  &0.708  &0.717  &0.623  &0.685  &\textbf{0.723}  \\
firearm    &0.582  &0.593 &0.544  &0.593  &0.600  &0.618  &0.664  &\textbf{0.693}  \\
couch    &0.643  &0.646 &0.628  &0.690  &0.706  &0.679  &0.723  &\textbf{0.749}  \\
table    &0.471  &0.536 &0.513  &0.564  &0.580  &0.547  &0.590  &\textbf{0.617}  \\
cellphone    &0.728 &0.702 &0.661  &0.732  &0.754  &0.738  &0.793  &\textbf{0.822}  \\
watercraft    &0.536 &\textbf{0.632} &0.513  &0.596  &0.610  &0.552  &0.606  &0.626  \\
\hline
\end{tabular}
\centerline{(b) Per-categoty results}
\end{minipage}
\caption{\label{fig:fig3}The overall (a) and per-category (b) multi-view reconstruction comparison by 3D-R2N2, ours and reference values of the sing-view based model PTN and OGN. Except for the watercraft, our method performs consistently the best in each category.}
\end{figure*}
%----------------------------------------------------------------------------------
\subsection{Implementation Details}
\label{section:section3.1}
Our model is trained and tested under the Pytorch framework, accelerated by a GPU (NVIDIA GTX 1080Ti). We
use the dataset from ~\cite{yan2016perspective}, which is based on the ShapeNetCore ~\cite{wu20153d}. Each model is represented as a 3D volume of $32\times32\times32$ from its canonical orientation, and images are rendered from $24$ azimuth angles with $30^{\circ}$ elevation angle. For each rendered image, we cropped and resized the centered region to $64\times64$ pixels with $3$ channels (RGB). We initialized all the weights using Xavier ~\cite{glorot2010understanding} and update the weights by using ADAM solver with batchsize 16, epoch 200, $\lambda_{vox}=\lambda_{proj}=0.5$.

\subsection{Evaluation on Reconstruction Performance}
\label{section:section3.2}

We compare our method with PTN (Perspective Transformer Network~\cite{yan2016perspective}), OGN (Octree Generating Networks~\cite{Tatarchenko2016Octree}) and 3D-R2N2 (proposed by Choy et al.~\shortcite{choy20163d}). PTN uses an encoder-decoder model to make a 3D volume prediction trained with a combined loss of both projection supervision and volume supervision. OGN generates volumetric 3D outputs in a compute- and memory-efficient manner by using an octree representation. To evaluate the multi-view performance, we also compare to 3D-R2N2, which performs both single- and multi-view 3D reconstruction using a 3D recurrent network. We trained and tested our network using 13 categories with train/test data split used by 3D-R2N2's authors, which is adopted by OGN's author as well. For a fair comparison, we followed 3D-R2N2's setting and used 5 random views along the view circle to evaluate our Recurrent Encoder-Decoder model. For PTN, We re-trained the model for multi-category reconstruction using the code released by the authors, since they originally trained their model only on chair category. In the test stage, we compute voxel IoU (\ref{eqn1} with threshold 0.4 as the evaluation metric.

%----------------------------------------------------------------------------------
\textbf{Overall results.} In Figure \ref{fig:fig3} (a), we plot the trend of mean reconstruction IoU by the compared methods. It can be seen that our method performs better than the baseline volumetric reconstruction methods PTN and OGN when using only single view and outperform it a large margin with an increasing number of views. It proves the ability of our model predicting a reasonable reconstruction result by using only a single image. Compared to \emph{\textbf{3D-R2N2}}, we get a significantly better reconstruction performance over the number of views \hb{, as confirmed by a two-sided t-test (p-value$<$0.01).} The reason is perhaps that our Recurrent Encoder-Decoder extracts more discriminative features and aggregates the information from different views at a deeper level.

\textbf{Per-category results.} We also examine the reconstruction performance of the compared methods on 13 categories as shown in the table of Figure \ref{fig:fig3} (b). Our model leads to higher IoUs with an increasing number of views and performs the best when using 5 views. Besides, we observe that our model does consistently better in single-view reconstruction than PTN and 3D-R2N2 as well. This may be benefited from our volume-projection guidance.

\textbf{Qualitative  results.} The examples of reconstruction results shown in Figure \ref{fig:fig4} qualitatively show that our model can generally make a reasonable prediction of a 3D object on a global shape even from a single view and succeed to optimize the local details that 3D-R2N2 fails (pointed out by the red circles) when using more information from different views.

%----------------------------------------------------------------------------------
\begin{figure}[htbp]
  \begin{minipage}{0.496\linewidth}
  \centerline{\includegraphics[width=1\textwidth]{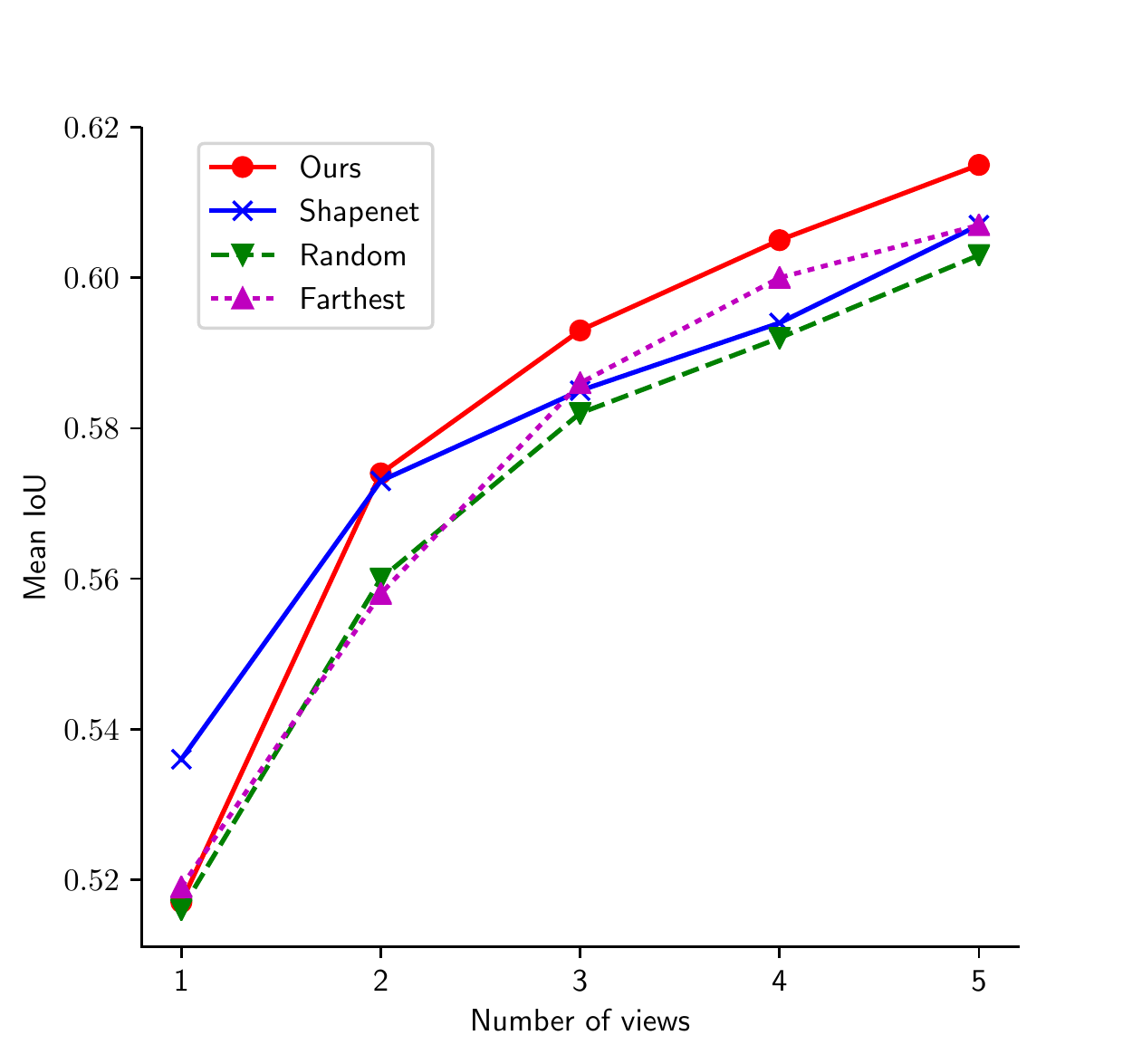}}
  \centerline{(a)}
  \end{minipage}
  \begin{minipage}{0.496\linewidth}
  \centerline{\includegraphics[width=1\textwidth]{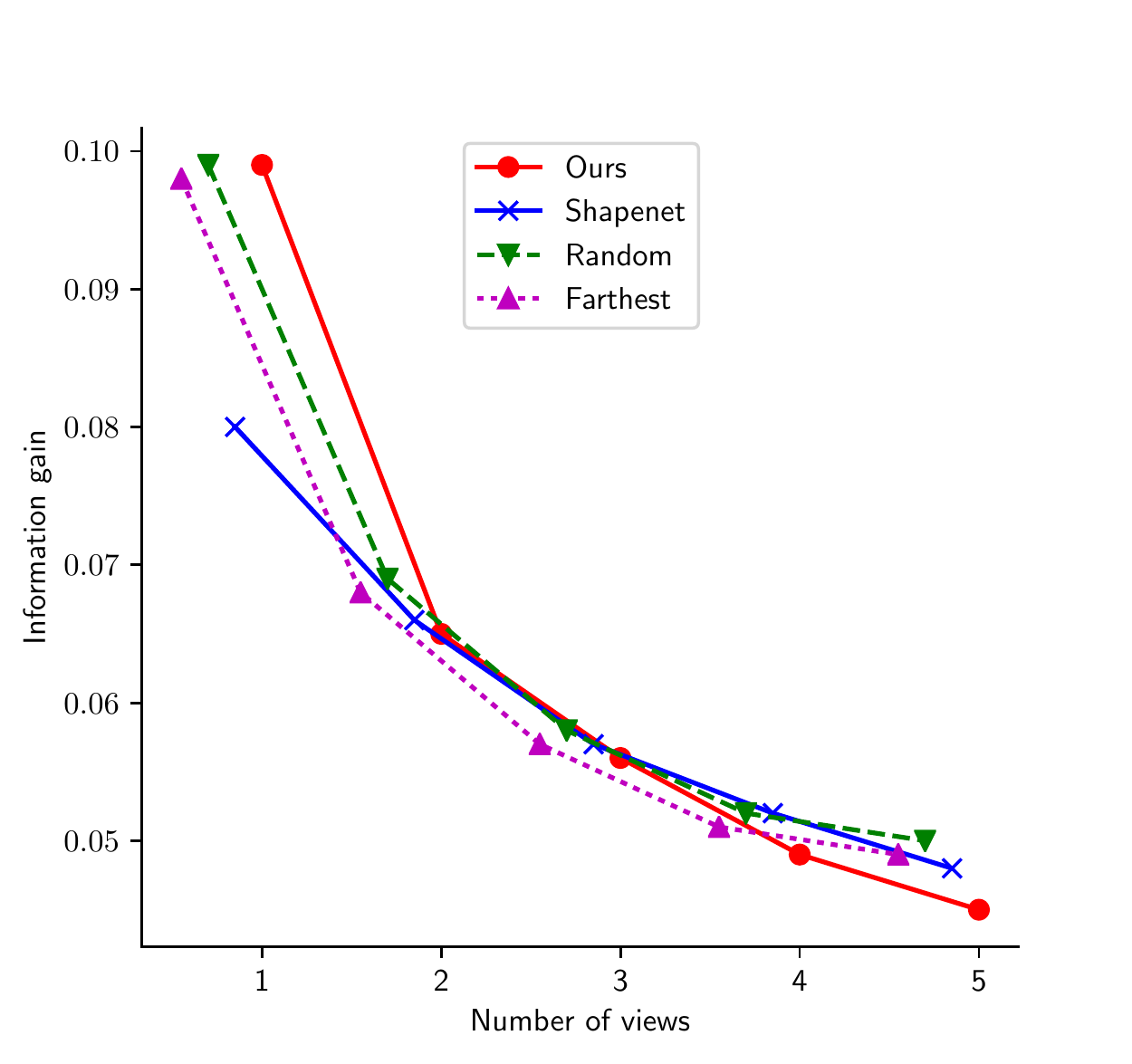}}
  \centerline{(b)}
  \end{minipage}
  \caption{ \label{fig:fig5}
    The view prediction comparison against the baselines (Random and Farthest) and ShapeNet. (a): IoU values (the higher and the better) over the number of views. (b): Information gain (the lower and the better) as the decrease of Shannon Entropy.}
\end{figure}
%----------------------------------------------------------------------------------
\subsection{Evaluation on Information Gain}

%----------------------------------------------------------------------------------
\begin{figure*}[tbp]
  \begin{minipage}{0.085\linewidth}
  Input Views
  \end{minipage}
  \begin{minipage}{0.085\linewidth}
  \centerline{\includegraphics[width=1\textwidth, height=1\textwidth]{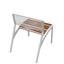}}
  \end{minipage}
  \begin{minipage}{0.085\linewidth}
  \centerline{\includegraphics[width=1\textwidth, height=1\textwidth]{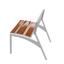}}
  \end{minipage}
  \begin{minipage}{0.085\linewidth}
  \centerline{\includegraphics[width=1\textwidth, height=1\textwidth]{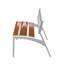}}
  \end{minipage}
  \begin{minipage}{0.085\linewidth}
  \centerline{\includegraphics[width=1\textwidth, height=1\textwidth]{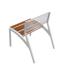}}
  \end{minipage}
  \begin{minipage}{0.085\linewidth}
  \centerline{\includegraphics[width=1\textwidth, height=1\textwidth]{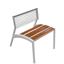}}
  \end{minipage}
  \hfill
  \begin{minipage}{0.085\linewidth}
  \centerline{\includegraphics[width=1\textwidth, height=1\textwidth]{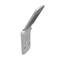}}
  \end{minipage}
  \begin{minipage}{0.085\linewidth}
  \centerline{\includegraphics[width=1\textwidth, height=1\textwidth]{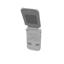}}
  \end{minipage}
  \begin{minipage}{0.085\linewidth}
  \centerline{\includegraphics[width=1\textwidth, height=1\textwidth]{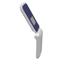}}
  \end{minipage}
  \begin{minipage}{0.085\linewidth}
  \centerline{\includegraphics[width=1\textwidth, height=1\textwidth]{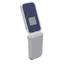}}
  \end{minipage}
  \begin{minipage}{0.085\linewidth}
  \centerline{\includegraphics[width=1\textwidth, height=1\textwidth]{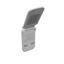}}
  \end{minipage}

  \begin{minipage}{0.085\linewidth}
  3D-R2N2
  \end{minipage}
  \begin{minipage}{0.085\linewidth}
  \centerline{\includegraphics[width=1\textwidth, height=1\textwidth]{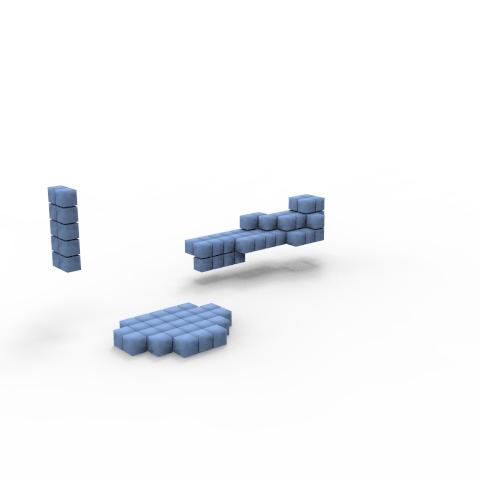}}
  \end{minipage}
  \begin{minipage}{0.085\linewidth}
  \centerline{\includegraphics[width=1\textwidth, height=1\textwidth]{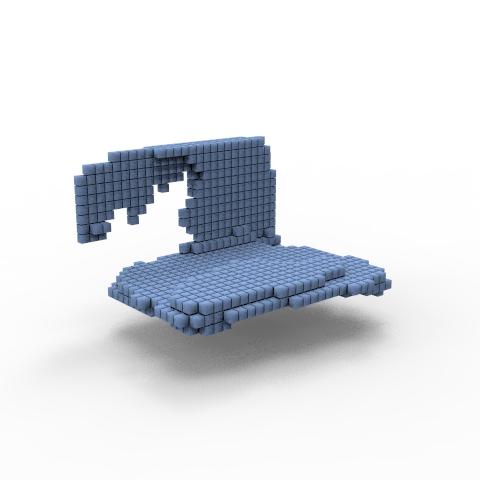}}
  \end{minipage}
  \begin{minipage}{0.085\linewidth}
  \centerline{\includegraphics[width=1\textwidth, height=1\textwidth]{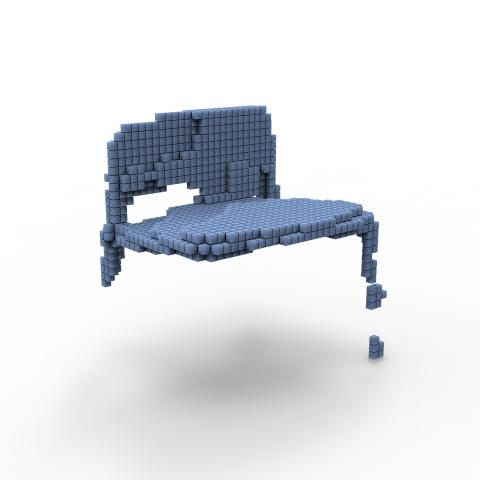}}
  \end{minipage}
  \begin{minipage}{0.085\linewidth}
  \centerline{\includegraphics[width=1\textwidth, height=1\textwidth]{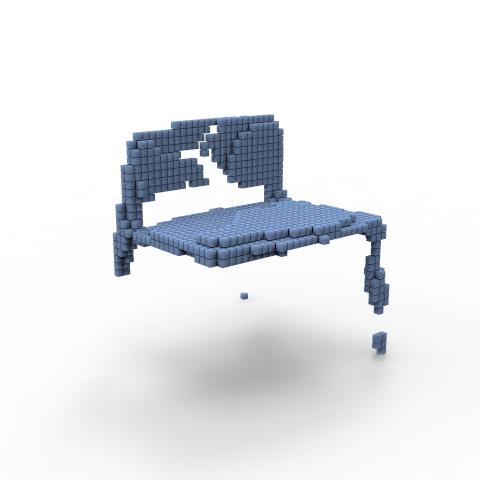}}
  \end{minipage}
  \begin{minipage}{0.085\linewidth}
  \centerline{\includegraphics[width=1\textwidth, height=1\textwidth]{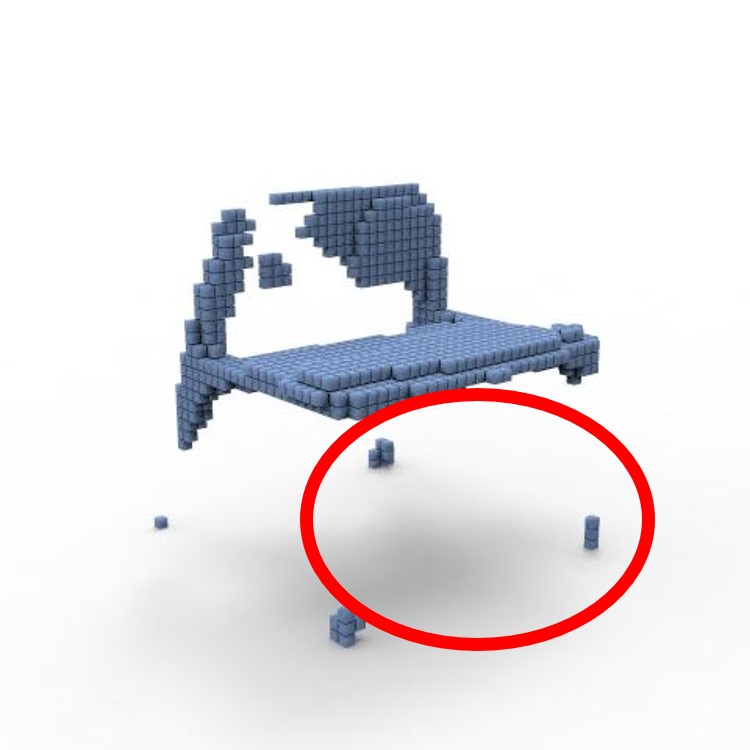}}
  \end{minipage}
  \hfill
  \begin{minipage}{0.085\linewidth}
  \centerline{\includegraphics[width=1\textwidth, height=1\textwidth]{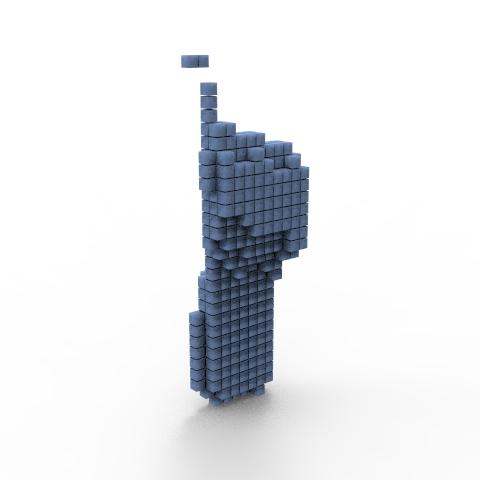}}
  \end{minipage}
  \begin{minipage}{0.085\linewidth}
  \centerline{\includegraphics[width=1\textwidth, height=1\textwidth]{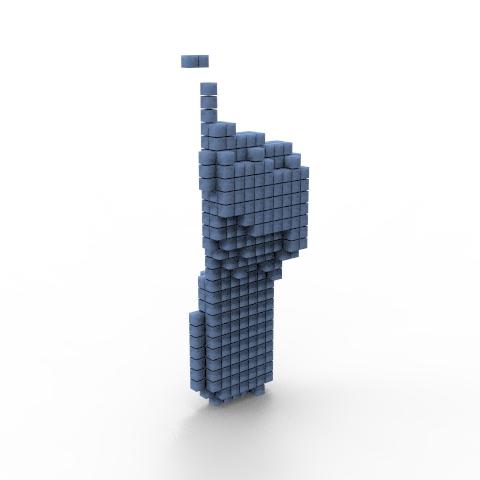}}
  \end{minipage}
  \begin{minipage}{0.085\linewidth}
  \centerline{\includegraphics[width=1\textwidth, height=1\textwidth]{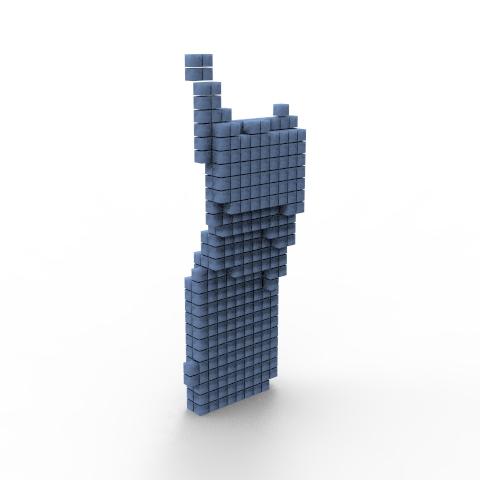}}
  \end{minipage}
  \begin{minipage}{0.085\linewidth}
  \centerline{\includegraphics[width=1\textwidth, height=1\textwidth]{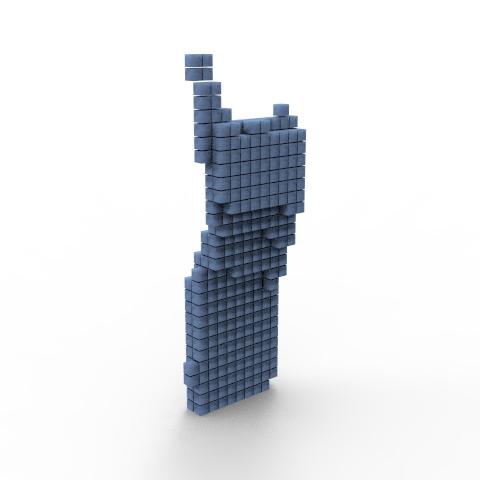}}
  \end{minipage}
  \begin{minipage}{0.085\linewidth}
  \centerline{\includegraphics[width=1\textwidth, height=1\textwidth]{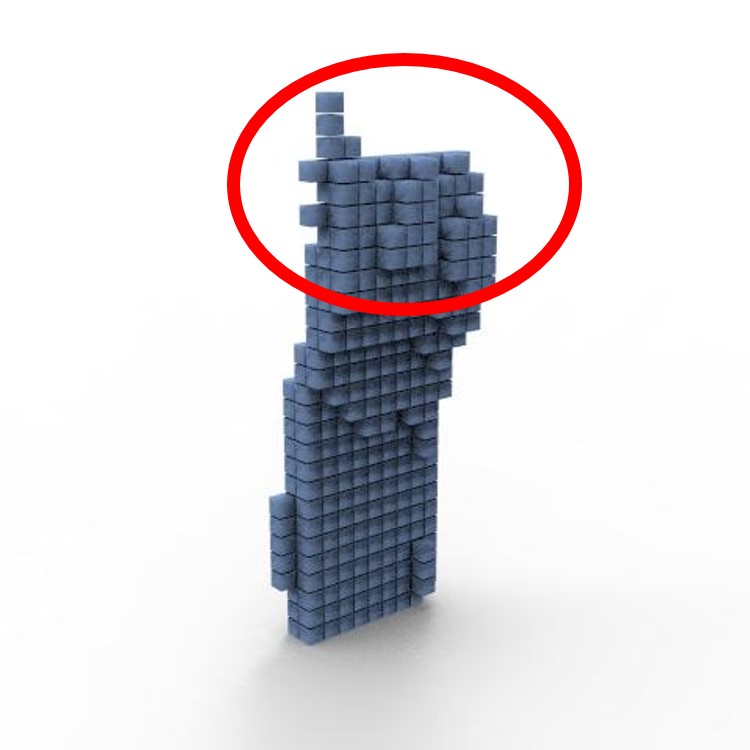}}
  \end{minipage}

  \begin{minipage}{0.085\linewidth}
  Ours
  \end{minipage}
  \begin{minipage}{0.085\linewidth}
  \centerline{\includegraphics[width=1\textwidth, height=1\textwidth]{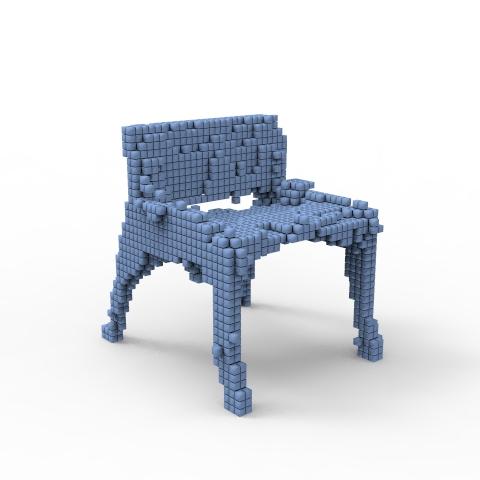}}
  \end{minipage}
  \begin{minipage}{0.085\linewidth}
  \centerline{\includegraphics[width=1\textwidth, height=1\textwidth]{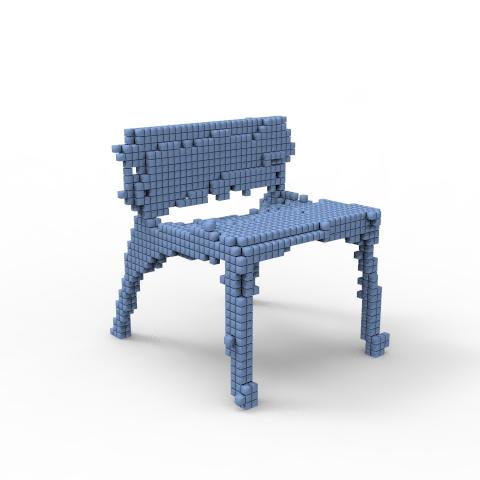}}
  \end{minipage}
  \begin{minipage}{0.085\linewidth}
  \centerline{\includegraphics[width=1\textwidth, height=1\textwidth]{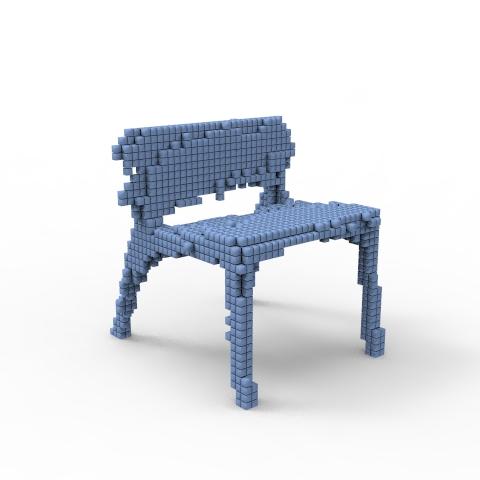}}
  \end{minipage}
  \begin{minipage}{0.085\linewidth}
  \centerline{\includegraphics[width=1\textwidth, height=1\textwidth]{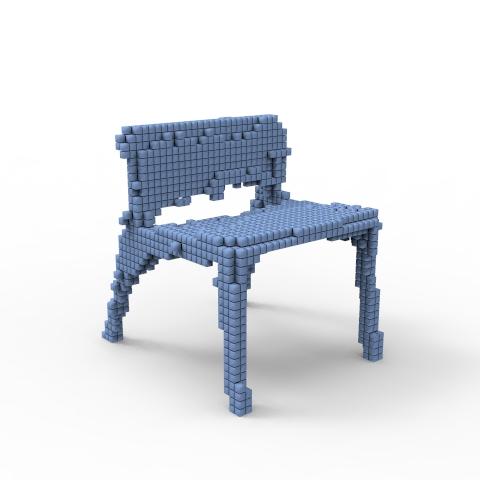}}
  \end{minipage}
  \begin{minipage}{0.085\linewidth}
  \centerline{\includegraphics[width=1\textwidth, height=1\textwidth]{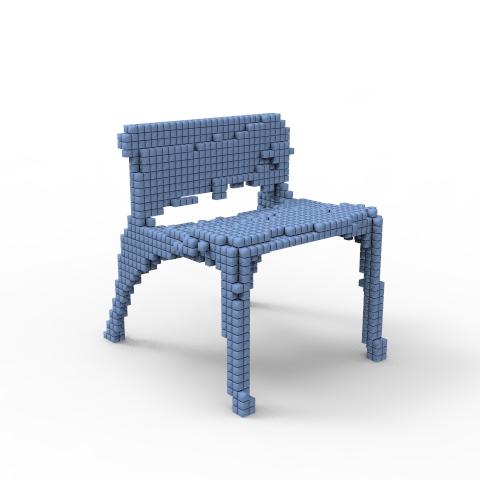}}
  \end{minipage}
  \hfill
  \begin{minipage}{0.085\linewidth}
  \centerline{\includegraphics[width=1\textwidth, height=1\textwidth]{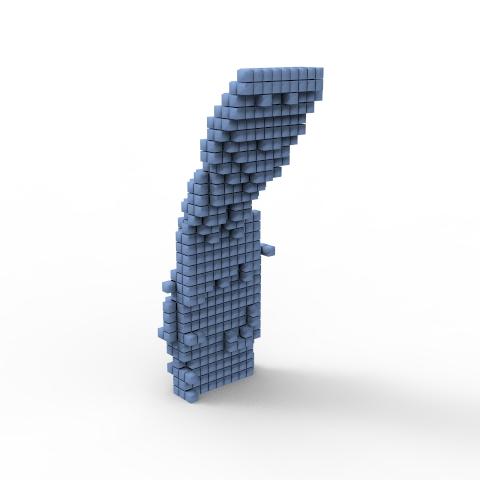}}
  \end{minipage}
  \begin{minipage}{0.085\linewidth}
  \centerline{\includegraphics[width=1\textwidth, height=1\textwidth]{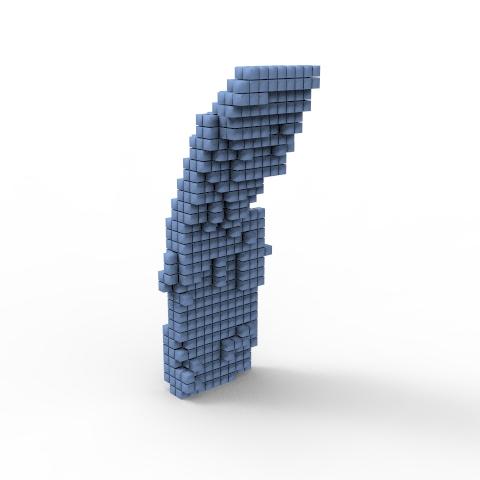}}
  \end{minipage}
  \begin{minipage}{0.085\linewidth}
  \centerline{\includegraphics[width=1\textwidth, height=1\textwidth]{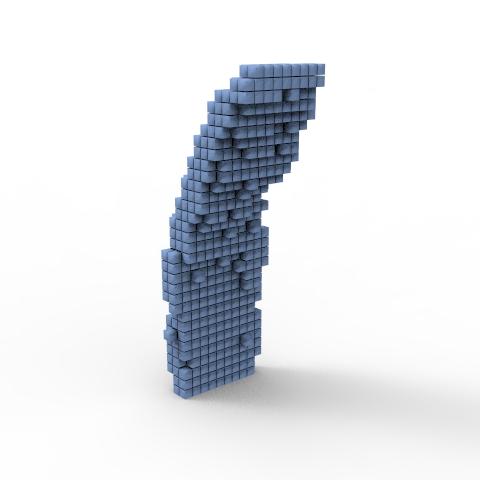}}
  \end{minipage}
  \begin{minipage}{0.085\linewidth}
  \centerline{\includegraphics[width=1\textwidth, height=1\textwidth]{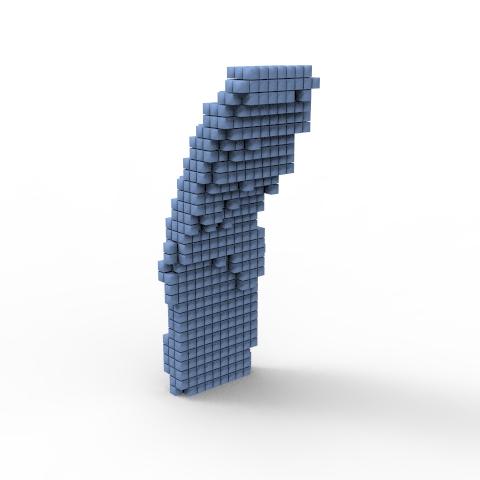}}
  \end{minipage}
  \begin{minipage}{0.085\linewidth}
  \centerline{\includegraphics[width=1\textwidth, height=1\textwidth]{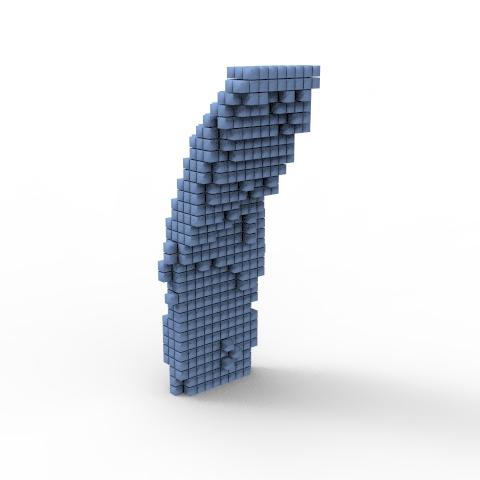}}
  \end{minipage}

  \begin{minipage}{0.085\linewidth}
  Input Views
  \end{minipage}
  \begin{minipage}{0.085\linewidth}
  \centerline{\includegraphics[width=1\textwidth, height=1\textwidth]{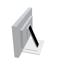}}
  \end{minipage}
  \begin{minipage}{0.085\linewidth}
  \centerline{\includegraphics[width=1\textwidth, height=1\textwidth]{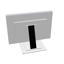}}
  \end{minipage}
  \begin{minipage}{0.085\linewidth}
  \centerline{\includegraphics[width=1\textwidth, height=1\textwidth]{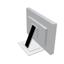}}
  \end{minipage}
  \begin{minipage}{0.085\linewidth}
  \centerline{\includegraphics[width=1\textwidth, height=1\textwidth]{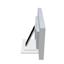}}
  \end{minipage}
  \begin{minipage}{0.085\linewidth}
  \centerline{\includegraphics[width=1\textwidth, height=1\textwidth]{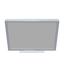}}
  \end{minipage}
  \hfill
  \begin{minipage}{0.085\linewidth}
  \centerline{\includegraphics[width=1\textwidth, height=1\textwidth]{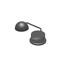}}
  \end{minipage}
  \begin{minipage}{0.085\linewidth}
  \centerline{\includegraphics[width=1\textwidth, height=1\textwidth]{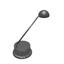}}
  \end{minipage}
  \begin{minipage}{0.085\linewidth}
  \centerline{\includegraphics[width=1\textwidth, height=1\textwidth]{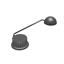}}
  \end{minipage}
  \begin{minipage}{0.085\linewidth}
  \centerline{\includegraphics[width=1\textwidth, height=1\textwidth]{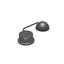}}
  \end{minipage}
  \begin{minipage}{0.085\linewidth}
  \centerline{\includegraphics[width=1\textwidth, height=1\textwidth]{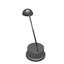}}
  \end{minipage}

  \begin{minipage}{0.085\linewidth}
  3D-R2N2
  \end{minipage}
  \begin{minipage}{0.085\linewidth}
  \centerline{\includegraphics[width=1\textwidth, height=1\textwidth]{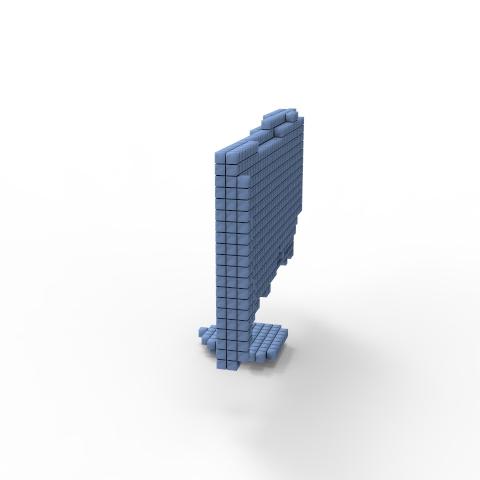}}
  \end{minipage}
  \begin{minipage}{0.085\linewidth}
  \centerline{\includegraphics[width=1\textwidth, height=1\textwidth]{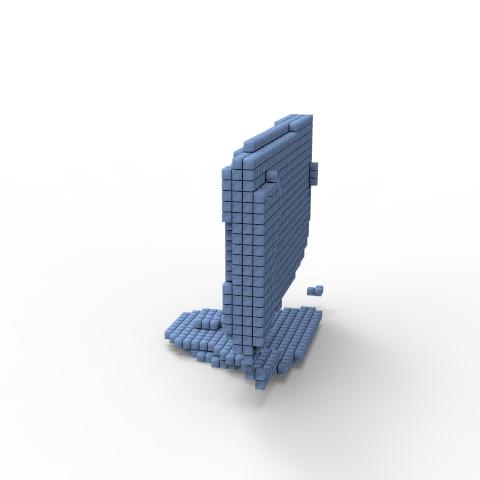}}
  \end{minipage}
  \begin{minipage}{0.085\linewidth}
  \centerline{\includegraphics[width=1\textwidth, height=1\textwidth]{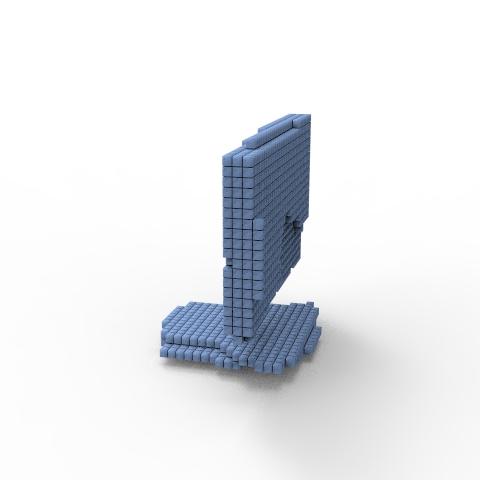}}
  \end{minipage}
  \begin{minipage}{0.085\linewidth}
  \centerline{\includegraphics[width=1\textwidth, height=1\textwidth]{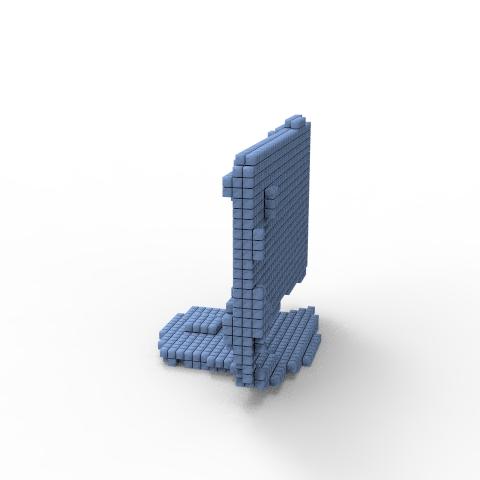}}
  \end{minipage}
  \begin{minipage}{0.085\linewidth}
  \centerline{\includegraphics[width=1\textwidth, height=1\textwidth]{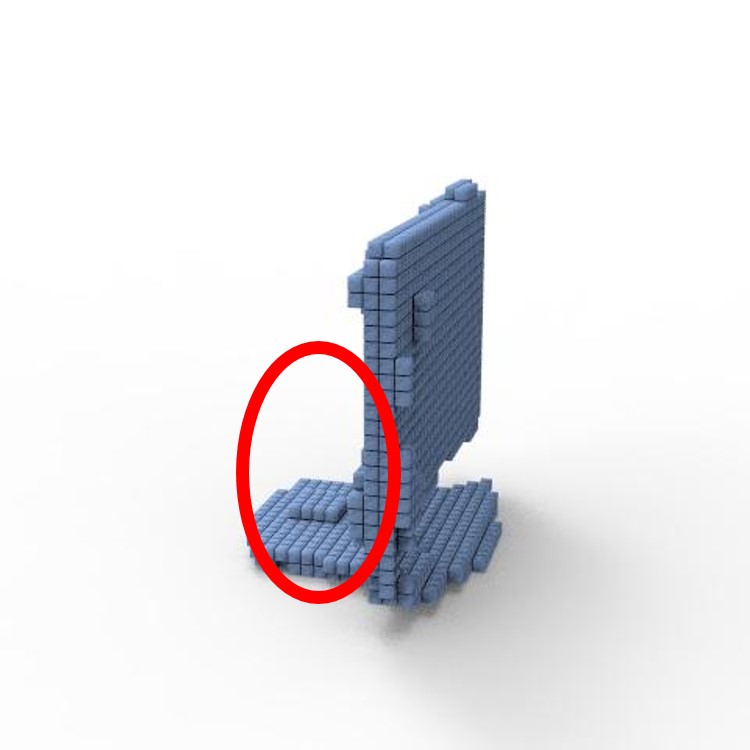}}
  \end{minipage}
  \hfill
  \begin{minipage}{0.085\linewidth}
  \centerline{\includegraphics[width=1\textwidth, height=1\textwidth]{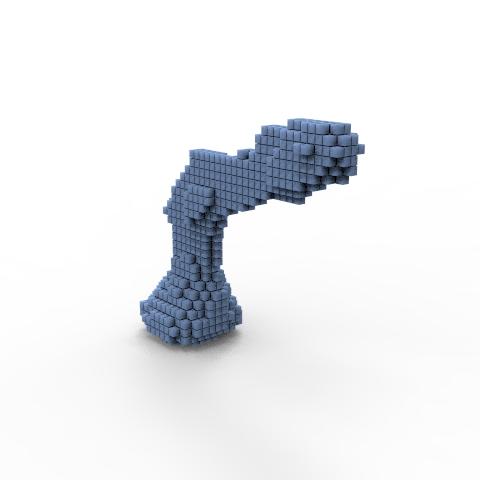}}
  \end{minipage}
  \begin{minipage}{0.08\linewidth}
  \centerline{\includegraphics[width=1\textwidth, height=1\textwidth]{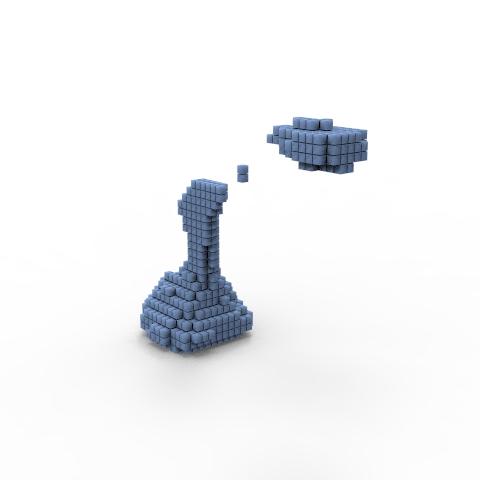}}
  \end{minipage}
  \begin{minipage}{0.085\linewidth}
  \centerline{\includegraphics[width=1\textwidth, height=1\textwidth]{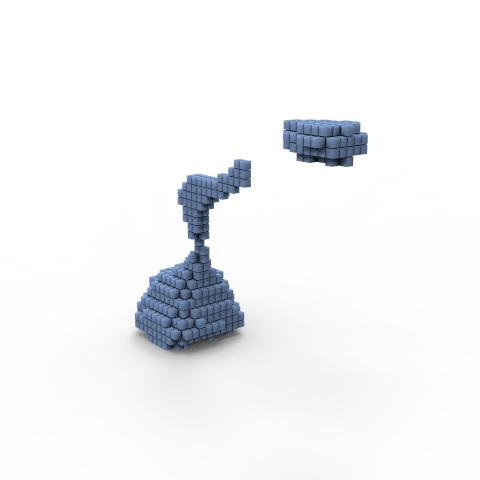}}
  \end{minipage}
  \begin{minipage}{0.085\linewidth}
  \centerline{\includegraphics[width=1\textwidth, height=1\textwidth]{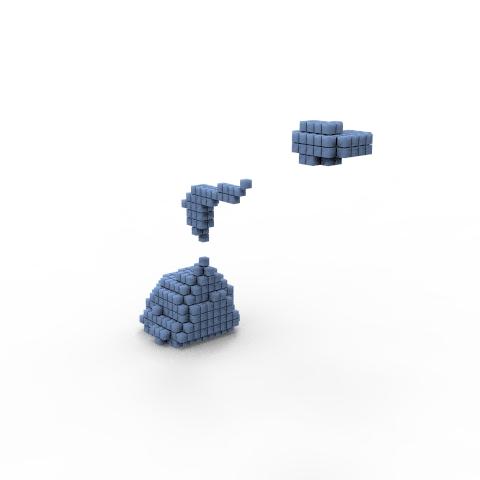}}
  \end{minipage}
  \begin{minipage}{0.085\linewidth}
  \centerline{\includegraphics[width=1\textwidth, height=1\textwidth]{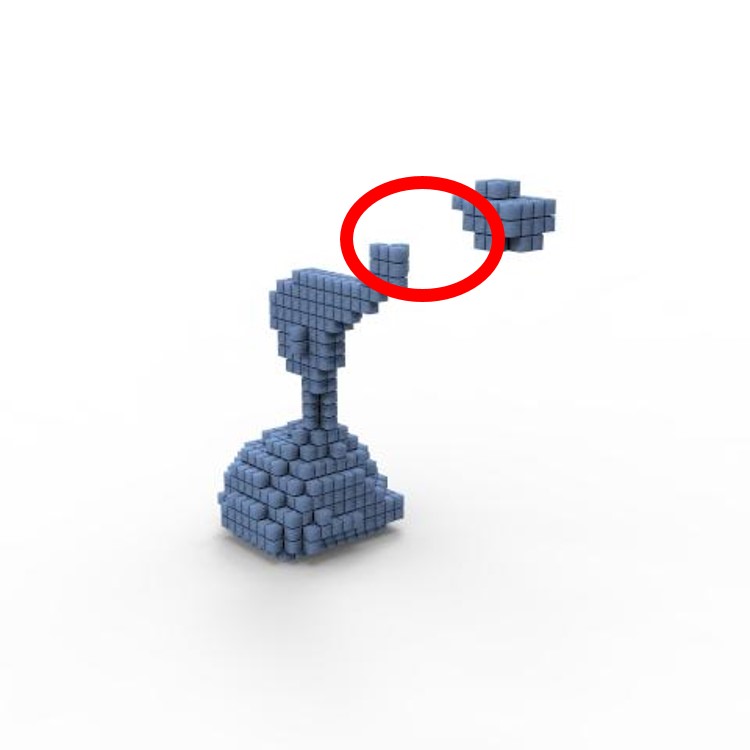}}
  \end{minipage}

  \begin{minipage}{0.085\linewidth}
  Ours
  \end{minipage}
  \begin{minipage}{0.085\linewidth}
  \centerline{\includegraphics[width=1\textwidth, height=1\textwidth]{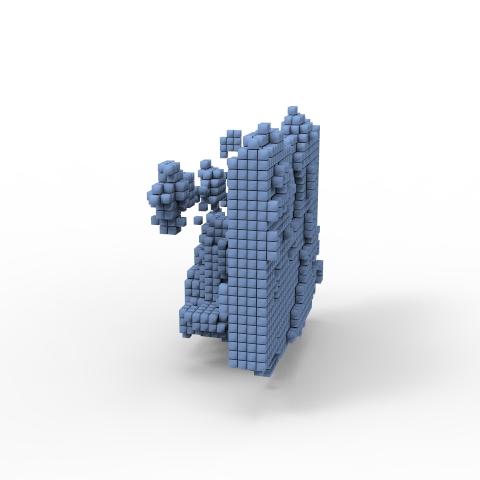}}
  \end{minipage}
  \begin{minipage}{0.085\linewidth}
  \centerline{\includegraphics[width=1\textwidth, height=1\textwidth]{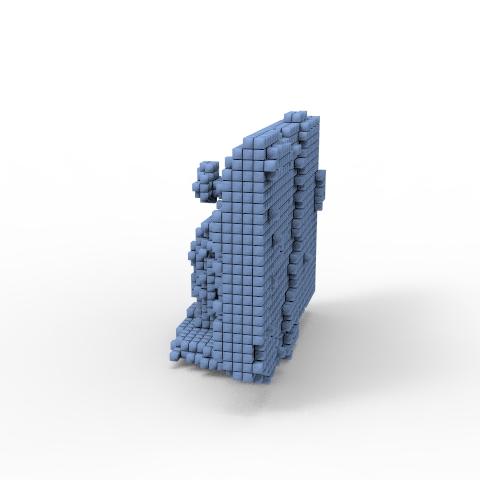}}
  \end{minipage}
  \begin{minipage}{0.085\linewidth}
  \centerline{\includegraphics[width=1\textwidth, height=1\textwidth]{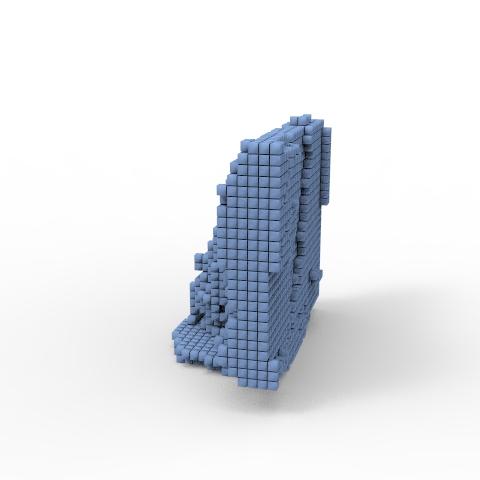}}
  \end{minipage}
  \begin{minipage}{0.085\linewidth}
  \centerline{\includegraphics[width=1\textwidth, height=1\textwidth]{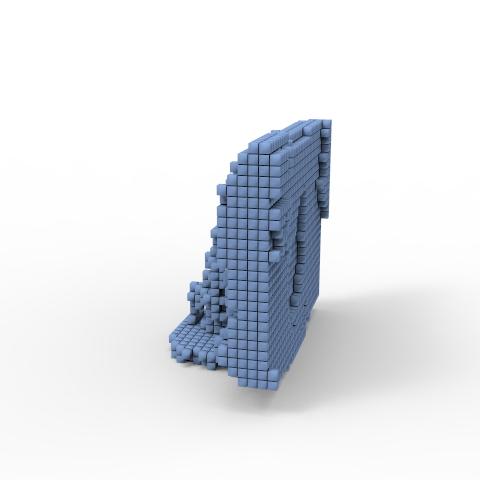}}
  \end{minipage}
  \begin{minipage}{0.085\linewidth}
  \centerline{\includegraphics[width=1\textwidth, height=1\textwidth]{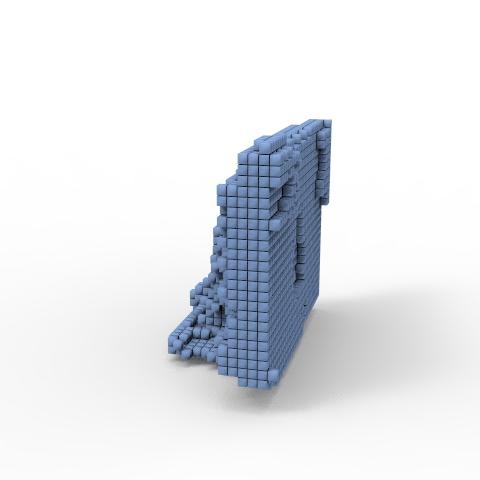}}
  \end{minipage}
  \hfill
  \begin{minipage}{0.085\linewidth}
  \centerline{\includegraphics[width=1\textwidth, height=1\textwidth]{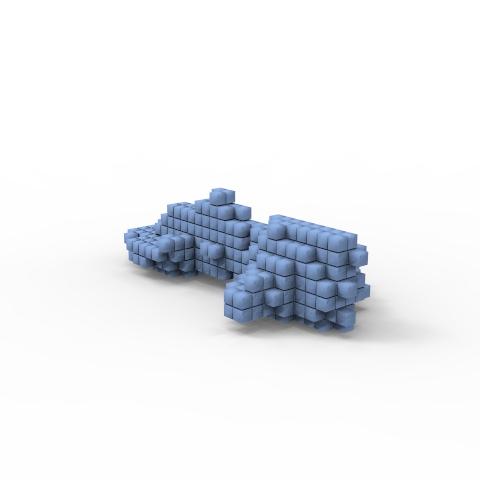}}
  \end{minipage}
  \begin{minipage}{0.085\linewidth}
  \centerline{\includegraphics[width=1\textwidth, height=1\textwidth]{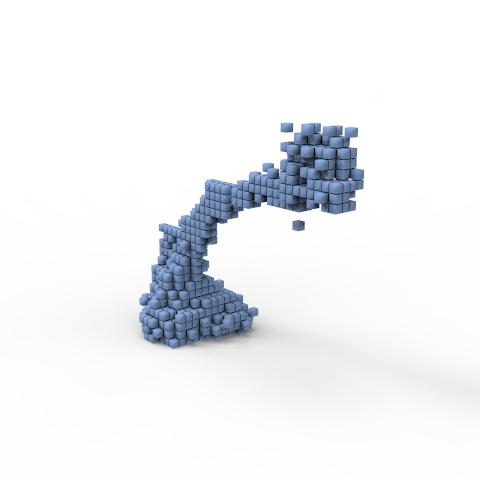}}
  \end{minipage}
  \begin{minipage}{0.085\linewidth}
  \centerline{\includegraphics[width=1\textwidth, height=1\textwidth]{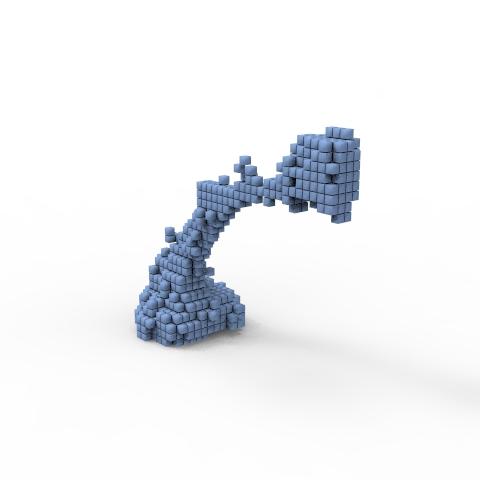}}
  \end{minipage}
  \begin{minipage}{0.085\linewidth}
  \centerline{\includegraphics[width=1\textwidth, height=1\textwidth]{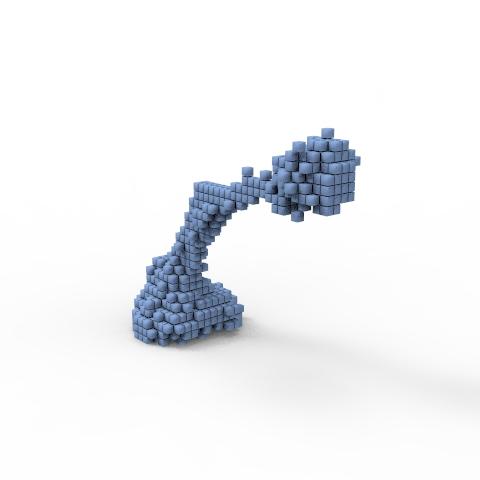}}
  \end{minipage}
  \begin{minipage}{0.085\linewidth}
  \centerline{\includegraphics[width=1\textwidth, height=1\textwidth]{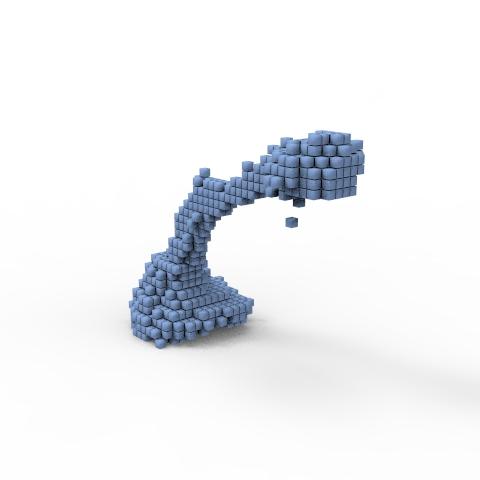}}
  \end{minipage}
  \vspace{1ex}
  \caption{ \label{fig:fig4}
           Qualitative results of reconstruction samples for example view sequences. 3D-R2N2 generally fails in the categories with much higer variation (eg. the lamp in the bottom right corner) while our model does better in feature extraction and view aggregation, leading to a more accurate reconstruction.}
\end{figure*}
%----------------------------------------------------------------------------------

\label{section:section3.3}
To evaluate the performance of view prediction, we compared our View planner against two baselines and an alternative method. The two baselines consist of a random planners that selects random view as the next view and a farthest planners that selects the view which is the farthest away from previous views in the viewing circle around the targeting object. The alternative methods is the NBV technique proposed in ShapeNet ~\cite{wu20153d} which estimates the information gain of a view from 3D volume. We train and test our active reconstruction model using the chair models in ShapeNet and rendering images under the train/test data split used by PTN's authors ~\cite{yan2016perspective}. We set $\lambda_{v}=10$, $\lambda_{p}=10$, $\lambda_{m}=0.04$. For comparison, we respectively feed the image sequence predicted by these strategies into the pre-trained Recurrent Encoder-Decoder to reconstruct the 3D volume.

We plot the IoU values and the decrease of Shannon Entropy over the number of views in Figure \ref{fig:fig5}. Compared to other methods, our model not only attains more information but also gets a more accurate results, showing that our model is able to predict a both informative and discriminative view sequence for more accurate reconstruction results.

\subsection{Network Structures Comparison}
\label{section:section3.4}
To demonstrate that our Recurrent Encoder-Decoder extracts a more discriminative feature from an image sequence and gets a better reconstruction performance, we compare four kinds of network architectures under different combinations: fully convolutional Encoder-Decoder with a 3D RNN (2E-R-3D), Recurrent 2D Encoder with a 3D CNN-based decoder (\emph{\textbf{R2E-3D}}), 2D CNN-based Encoder with a Recurrent 3D Decoder (\emph{\textbf{2E-R3D}}), and our Recurrent 2D Encoder with Recurrent 3D Decoder (\emph{\textbf{R2E-R3D}}). We trained all these four models on \work{the} chair category using the rendered images from random views and ground truth 3D volume under the train/test data split of ShapeNet database used by PTN's authors. For comparison, we utilize 5 random views to reconstruct the 3D volume and show in Table \ref{table:table 2} the results of MSE loss and IoU values. The results show that our \emph{\textbf{R2E-R3D}} architecture \hb{performs the best on both training losses and testing IoU values}. Using \emph{\textbf{R2E-R3D}} model, we can achieve the best reconstruction performance against the other settings, which validates our model superior in view-based reconstruction task.

%%-----------------------------------------------------------------------
\begin{table}[htbp]
\begin{center}
\begin{tabular}{| c | c | c | c | c |}
\hline
{Structure} &{2E-R-3D} &{R2E-3D} &{2E-R3D} &{R2E-R3D}\\
\hline
Encoder    &2D Enc  &R-2D  &2D Enc  &R-2D\\
Decoder     &3D Dec  &3D Dec  &R-3D  &R-3D\\
\hline
Loss  &0.0241  &0.021  &0.014  &\textbf{0.012}\\
IoU   &0.6285  &0.704   &0.785  &\textbf{0.798}\\
IoU(test)	 &0.519  &0.542   &0.571  &\textbf{0.605}\\
\hline
\end{tabular}
\vspace{1ex}
\caption{The comparison of quantitative results under different variations on network structures (evaluated using 5 views).}
\label{table:table 2}
\end{center}
\end{table}
%-------------------------------------------------------------------------
\section{Conclusion}
In this paper, we have presented an learning-based model with active perception which unifies the guided information acquisition and multi-view object reconstruction. Under the guidance from both volume and projection, we jointly train the Recurrent Encoder-Decoder and View Planner for active object reconstruction. Experiments demonstrate that our model obtains more information and increases the reconstruction performance with an increasing number of views. \yuanbo{Our model only extracts the semantic features but ignores the correspondence of geometrical features from different camera viewpoints, leading to a slow growth when feeding in more than 5 views. In the future, we would utilize multi-modal features to optimize or jointly learn the object reconstruction and utilize more efficient data representations to increase the output resolution. Besides,} it is interesting to extend our approach to multi-object reconstruction by predicting the transformation of camera view from one object to another one.

\section*{Acknowledgments}
We thank the anonymous reviewers for the insightful and constructive comments. The work was partially funded by the Research Grants Council of HKSAR, China (Project No. CityU 11237116 and CityU 11300615), ACIM-SCM, the Hong Kong Scholars Program, and by NSFC grant from National Natural Science Foundation of China (NO. 91748104, 61632006, 61425002, U1708263).

%The

\appendix

\bibliographystyle{named}
\bibliography{ijcai18}

\end{document}